\documentclass[a4paper,11pt]{article}
\usepackage[margin=2cm]{geometry}
\usepackage{amsfonts}
\usepackage{amsmath}
\usepackage{mathptmx}
\usepackage{amssymb}
\usepackage{graphics}
\usepackage{graphicx}
\usepackage{times}
\usepackage[linktocpage,colorlinks=true,urlcolor=blue,citecolor=red,bookmarks=false]{hyperref}
\usepackage[font=small,bf]{caption}
\usepackage{morefloats}
\usepackage{lscape}
\usepackage{listings}

\newcommand{\reftocode}[1]{\texttt{#1}}

\usepackage{color}

\definecolor{mygreen}{rgb}{0,0.6,0}
\definecolor{mygray}{rgb}{0.5,0.5,0.5}
\definecolor{mymauve}{rgb}{0.58,0,0.82}

\DeclareMathAlphabet{\mathcal}{OMS}{cmsy}{m}{n}

\title{A Framework for the Volumetric Integration of Depth Images}
\author{
\makebox[.4\linewidth]{Victor Adrian Prisacariu*} \\ \url{victor@robots.ox.ac.uk}  \\ University of Oxford \and
\makebox[.4\linewidth]{Olaf~K\"{a}hler*} \\ \url{olaf@robots.ox.ac.uk} \\ University of Oxford \and
\makebox[.4\linewidth]{Ming Ming Cheng} \\ \url{cmm.thu@gmail.com} \\ Nankai University, China\and
\makebox[.4\linewidth]{Carl Yuheng Ren} \\ \url{carl@robots.ox.ac.uk} \\ University of Oxford \and
\makebox[.4\linewidth]{Julien Valentin} \\ \url{julien.valentin@eng.ox.ac.uk} \\ University of Oxford \and
\makebox[.4\linewidth]{Philip H.S. Torr} \\ \url{philip.torr@eng.ox.ac.uk} \\ University of Oxford \and
\makebox[.4\linewidth]{Ian D Reid} \\ \url{ian.reid@adelaide.edu.au} \\ University of Adelaide \and
\makebox[.4\linewidth]{David W Murray} \\ \url{dwm@robots.ox.ac.uk} \\ University of Oxford}

\begin{document}
\maketitle
\let\thefootnote\relax\footnotetext{* Olaf K\"{a}hler and Victor Adrian Prisacariu contributed equally to this work}

\tableofcontents

\section{Introduction} \label{s:infinitam_introduction}

Volumetric models have become a popular representation for 3D scenes in recent
years. While being used as early as~\cite{Curless96:VMB}, one of the
breakthroughs leading to their popularity was
KinectFusion~\cite{newcombe_ismar_2011,izadi_uist_2011}. The focus of
KinectFusion and a number of related works~\cite{ReconstructME_2012,kinfu_2011,Reitmayr13:KFusion}
is on 3D reconstruction using RGB-D sensors. However, monocular SLAM has since also
been tackled with very similar approaches~\cite{Newcombe11:DTA,Pradeep13:MFR}.
In the monocular case, various dense stereo techniques are used to create depth
images and then the same integration methods and volumetric representations are
used to fuse the information into a consistent 3D world model. Storing the
underlying truncated signed distance function (TSDF) volumetrically makes for
most of the simplicity and efficiency that can be achieved with GPU
implementations of these systems. However, this representation is also
memory-intensive and limits the applicability to small scale reconstructions.

Several avenues have been explored for overcoming this limitation. A first line
of works uses a moving volume~\cite{Roth12:MVK,Whelan13:RRT,kinfuls_2012} to
follow the camera while a sparse point cloud or mesh representation is computed
for the parts of the scene outside the active volume. In a second line of
research, the volumetric representation is split into a set of blocks aligned
with dominant planes~\cite{Henry13:PVS} or even reduced to a set of bump maps
along such planes~\cite{Thomas13:FSR}. A third category of approaches employs
an octree representation of the 3D volume~\cite{Zeng13:OFR,Chen13:SRV,
Steinbruecker14:V3D}. Finally, a hash lookup for a set of sparsely allocated
subblocks of the volume is used in~\cite{neissner_tog_2013}. Some of these
works, e.g.~\cite{kinfuls_2012,Chen13:SRV,neissner_tog_2013,Henry13:PVS}, also
provide methods for swapping data from the limited GPU memory to some larger
host memory to further expand the scale of the reconstructions.

With the aim of providing for a fast and flexible 3D reconstruction pipeline,
we propose a new, unifying framework called InfiniTAM. The core idea is that
individual steps like camera tracking, scene representation and integration of
new data can easily be replaced and adapted to the needs of the user.

Along with the framework we also provide a set of components for scalable
reconstruction: two implementations of camera trackers, based on RGB data and
on depth data, two representations of the 3D volumetric data, a dense volume
and one based on hashes of subblocks~\cite{neissner_tog_2013}, and an optional
module for swapping subblocks in and out of the typically limited GPU memory.
Given these components, a wide range of systems can be developed for specific
purposes, ranging from very efficient reconstruction of small scale volumes
with limited hardware resources up to full scale reconstruction of large-scale
scenes. While such systems can be tailored for the development of higher level
applications, the framework also allows users to focus on the development of
individual new components, reusing only parts of the framework.

Although most of the ideas used in our implementation have already been
presented in related works, there are a number of differences
and novelties that went into the engineering of our framework. For example in
Section~\ref{s:swapping} we define an engine for swapping data to and from the
GPU in a way that can deal with slow read and write accesses on the host,
e.g.\ on a disk. It also has a fixed maximum number of data transfers between
GPU and host to ensure an interactive online framework.
One general aim of the whole framework was to keep the implementation portable,
adaptable and simple. As we show in Section~\ref{s:results}, InfiniTAM has
minimal dependencies and natively builds on Linux, Mac OS and Windows platforms.

The remainder of this report and in particular Sections~\ref{s:itm_overview}
through~\ref{s:swapping} describe technical implementation details of the
InfiniTAM framework. These sections are aimed at closing the gap between the
theoretical description of volumetric depth map fusion and the actual software
implementation in our InfiniTAM package. Some advice on
compilation and practical usage of InfiniTAM is given in Section~\ref{s:results}
and some concluding remarks follow in Section~\ref{s:conclusions}.

\section{Architecture Overview}\label{s:itm_overview}
We first present an overview of the overall processing pipeline of our
framework and discuss the cross device implementation architecture. These
serve as the backbone of the framework throughout this report.

\subsection{Processing Pipeline}\label{s:pipeline}
\begin{figure}[tb]
\centering
\includegraphics[width=\linewidth]{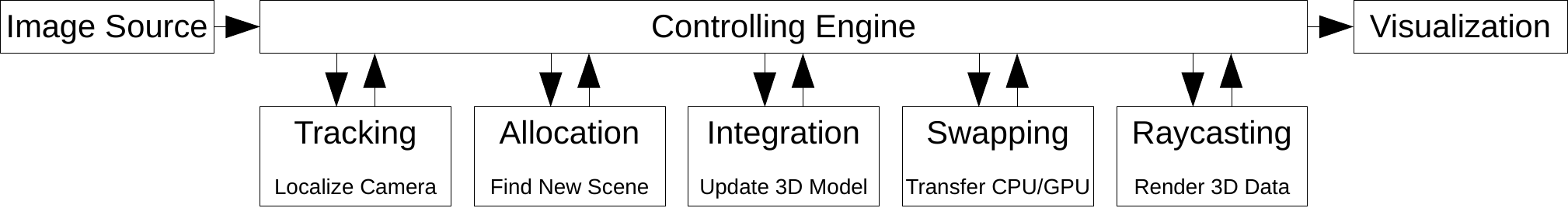}
\caption{InfiniTAM processing pipeline}
\label{fig:itm_pipeline}
\end{figure} 
The main processing steps are illustrated in Figure~\ref{fig:itm_pipeline}.
As with the typical and well known KinectFusion
pipeline~\cite{newcombe_ismar_2011}, there is a \textit{tracking} step for
localizing the camera, an \textit{integration} step for updating the 3D world
model and a \textit{raycasting} step for rendering the 3D data in preparation
of the next tracking step. To accommodate the additional processing requirements
for octrees~\cite{Steinbruecker14:V3D}, voxel block
hashing~\cite{neissner_tog_2013} or other data structures, an
\textit{allocation} step is included, where the respective data structures are
updated in preparation for integrating data from a new frame. Also an optional
\textit{swapping} step is included for transferring data between GPU and CPU.

A detailed discussion of each individual stage follows in
Section~\ref{s:itm_methodstages}. For now notice that our implementation
follows the chain-of-responsibility design pattern, which means that the
relevant data structures (e.g.\ \reftocode{ITMImage}) 
are passed between several processing engines
(e.g.\ \reftocode{ITMSceneReconstructionEngine}). The engines
are stateless and each engine is responsible for one specific aspect of
the overall processing pipeline. The state is passed on in objects containing
the processed information. Finally one central class (\reftocode{ITMMainEngine})
holds instances of all objects and engines and controls the flow of information.

\subsection{Cross Device Implementation Architecture} \label{s:itm_codearch}
\begin{figure}[tb]
\centering
\includegraphics[width=0.75\linewidth]{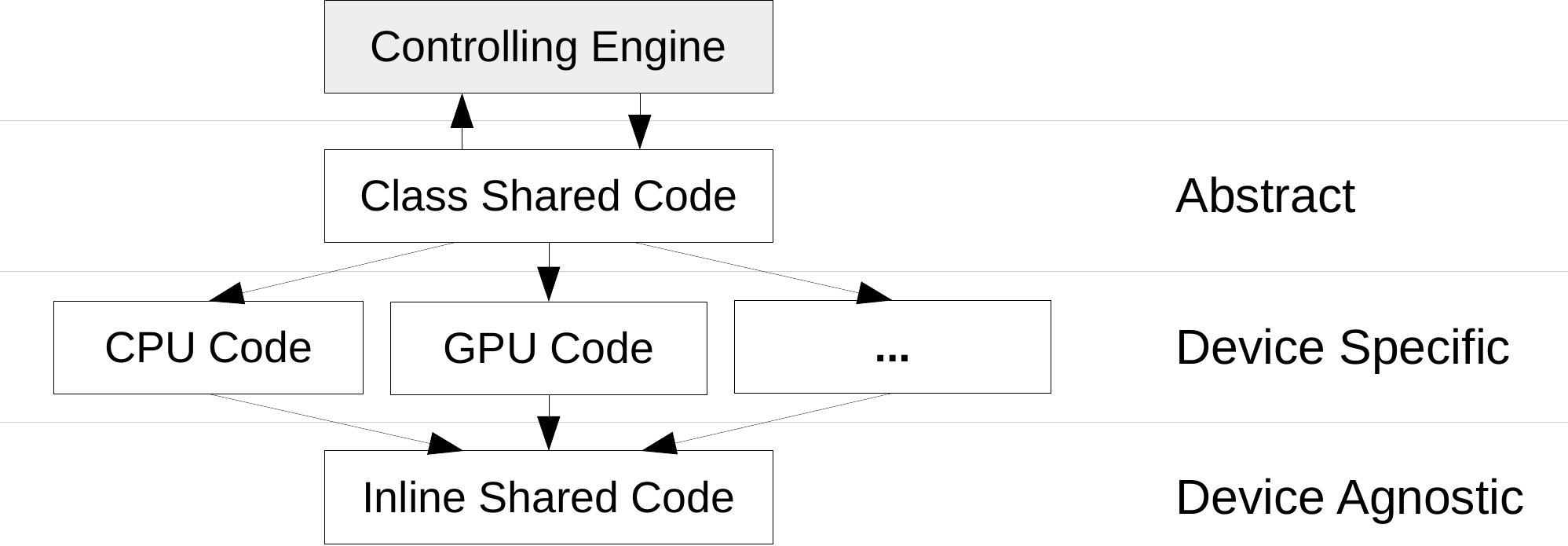}
\caption{InfiniTAM Cross Device Engine Architecture}
\label{fig:itm_enginestruct}
\end{figure} 

Each engine is further split into three layers as shown in
Figure~\ref{fig:itm_enginestruct}. The topmost, \textit{Abstract
Layer} is accessed by the library's main engine and is in general just an
blank interface, although some common code can be shared at this point. The
interface is implemented in the \textit{Device Specific Layer},
which will be very different depending on whether it runs on a CPU, a GPU,
on OpenMP or other hardware acceleration architectures. In the \textit{Device
Agnostic Layer} there may be some inline code that is called from the higher
layers and recompiled for the different architectures.

Considering the example of a tracking engine, 
the \textit{Abstract Layer} contains
code for the generic optimisation of an error function, the \textit{Device
Specific Layer} contains a loop or GPU kernel call to evaluate the error
function for all pixels in an image, and the \textit{Device Agnostic Layer}
contains a simple inline C-function to evaluate the error in a single pixel.

\section{Volumetric Representation}\label{s:itm_data_structures}
The volumetric representation is the central data structure within the
InfiniTAM framework. It is used to store a truncated signed distance function
that implicitly represents the 3D geometry by indicating for each point in
space how far in front or behind the scene surface it is.
Being truncated, only a small band of values around the current surface
estimate is actually being stored, while beyond the band the values are
set to a maximum and minimum value, respectively.

The data structure used for this representation crucially defines the memory
efficiency of the 3D scene model and the computational efficiency of
interacting with this model. We therefore keep it modular and interchangeable
in our framework and provide two implementations, namely for a fixed size dense
volume (\reftocode{ITMPlainVoxelArray}) and voxel block hashing
(\reftocode{ITMVoxelBlockHash}). Both are described in the following, but first
we briefly discuss how the representation is kept interchangeable throughout
the framework.

\subsection{Framework Flexibility}
Core data classes dealing with the volumetric signed distance function are
templated on (i) the type of voxel information stored, and (ii) the data
structure used for indexing and accessing the voxels. The relevant engine
classes are equally templated to provide the specific implementations
depending on the choosen data structures.

An example for a class representing the actual information at each voxel
is given in the following listing:
\begin{lstlisting}
struct ITMVoxel_s_rgb
{
	_CPU_AND_GPU_CODE_ static short SDF_initialValue() { return 32767; }
	_CPU_AND_GPU_CODE_ static float SDF_valueToFloat(float x)
	{ return (float)(x) / 32767.0f; }
	_CPU_AND_GPU_CODE_ static short SDF_floatToValue(float x)
	{ return (short)((x) * 32767.0f); }

	static const bool hasColorInformation = true;

	/** Value of the truncated signed distance transformation. */
	short sdf;
	/** Number of fused observations that make up @p sdf. */
	uchar w_depth;
	/** RGB colour information stored for this voxel. */
	Vector3u clr;
	/** Number of observations that made up @p clr. */
	uchar w_color;

	_CPU_AND_GPU_CODE_ ITMVoxel_s_rgb()
	{
		sdf = SDF_initialValue();
		w_depth = 0;
		clr = (uchar)0;
		w_color = 0;
	}
};
\end{lstlisting}
where the macro \reftocode{\_CPU\_AND\_GPU\_CODE\_} identifies methods and
functions that can be run both as host and as device code and is defined as:
\begin{lstlisting}
#if defined(__CUDACC__) && defined(__CUDA_ARCH__)
#define _CPU_AND_GPU_CODE_ __device__	// for CUDA device code
#else
#define _CPU_AND_GPU_CODE_ 
#endif
\end{lstlisting}
Alternative voxel types provided along with the current implementation are
based on floating point values instead of short, or they do not contain colour
information. Note that the member \reftocode{hasColorInformation} is used in
the provided integration methods to decide whether or not to gather colour
information, which has an impact on the processing speed accordingly.

Two examples for the index data structures that allow accessing the voxels are
given below. Again, note that the processing engines choose different
implementations according to the selected indexing class. For example the
allocation step for voxel block hashing has to modify a hash table, whereas
for a dense voxel array it can return immediately and does not have to do
anything.

\subsection{Dense Volumes}
We first discuss the naive way of using a dense volume of limited size, as
presented in the earlier works of~\cite{newcombe_ismar_2011,izadi_uist_2011,
Newcombe11:DTA,Pradeep13:MFR}. This is very well suited for understanding the
basic workings of the algorithms, it is trivial to parallelise on the GPU, and
it is sufficient for small scale reconstruction tasks.

In the InfiniTAM framework the class \reftocode{ITMPlainVoxelArray} is used to
represent dense volumes and a simplified listing of this class is given below:
\begin{lstlisting}
class ITMPlainVoxelArray
{
	public:
	struct ITMVoxelArrayInfo {
		/// Size in voxels
		Vector3i size;
		/// offset of the lower left front corner of the volume in voxels
		Vector3i offset;

		ITMVoxelArrayInfo(void)
		{
			size.x = size.y = size.z = 512;
			offset.x = -256;
			offset.y = -256;
			offset.z = 0;
		}
	};

	typedef ITMVoxelArrayInfo IndexData;

	private:
	IndexData indexData_host;

	public:
	ITMPlainVoxelArray(bool allocateGPU)
	...

	~ITMPlainVoxelArray(void)
	...

	/** Maximum number of total entries. */
	int getNumVoxelBlocks(void) { return 1; }
	int getVoxelBlockSize(void) { return indexData_host.size.x * indexData_host.size.y * indexData_host.size.z; }

	const Vector3i getVolumeSize(void) { return indexData_host.size; }

	const IndexData* getIndexData(void) const
	...
};
\end{lstlisting}
Note that the subtype \reftocode{IndexData} as well as the methods
\reftocode{getIndexData()}, \reftocode{getNumVoxelBlocks()} and
\reftocode{getVoxelBlockSize()} are used extensively within the processing
engines and on top of that the methods \reftocode{getNumVoxelBlocks()} and 
\reftocode{getVoxelBlockSize()} are used to determine the overall number of
voxels that have to be allocated.

Depending on the choice of voxel type, each voxel requires at least 3 bytes
and, depending on the available memory a total volume of about
$768\times{}768\times{}768$
voxels is typically the upper limit for this dense representation. To make the
most of this limited size, the initial camera is typically placed towards the
centre of the rear plane of this voxel cube, as identified by the
\reftocode{offset} in our implementation, so that there is maximum overlap
between the view frustum and the volume.

\subsection{Voxel Block Hashing}
The key idea for scaling the SDF based representation to larger 3D environments is
to drop the empty voxels from outside the truncation band and to represent only
the relevant parts of the volume. In~\cite{neissner_tog_2013} this is achieved
using a hash lookup of subblocks of the volume. Our framework provides an
implementation of this method that we explain in the following.

Voxels are grouped in blocks of predefined size (currently $8\times8\times8$
voxels). All the voxel blocks are stored in a contiguous array, referred
henceforth as the \textit{voxel block array} or VBA. In the current
implementation this has a defined size of $2^{18}$ elements. 

To handle the voxel blocks we further need:
\begin{itemize}
\item Hash Table and Hashing Function: Enable fast access to voxel blocks in the voxel block array -- details in Subsection \ref{ss:itm_hashtable}.
\item Hash Table Operations: Insertion, retrieval and deletion of voxel blocks -- details in Subsection \ref{ss:itm_hashopers}.
\end{itemize}

\subsubsection{Hash Table and Hashing Function} \label{ss:itm_hashtable}
\begin{figure}[tb]
\centering
\includegraphics[width=0.85\linewidth]{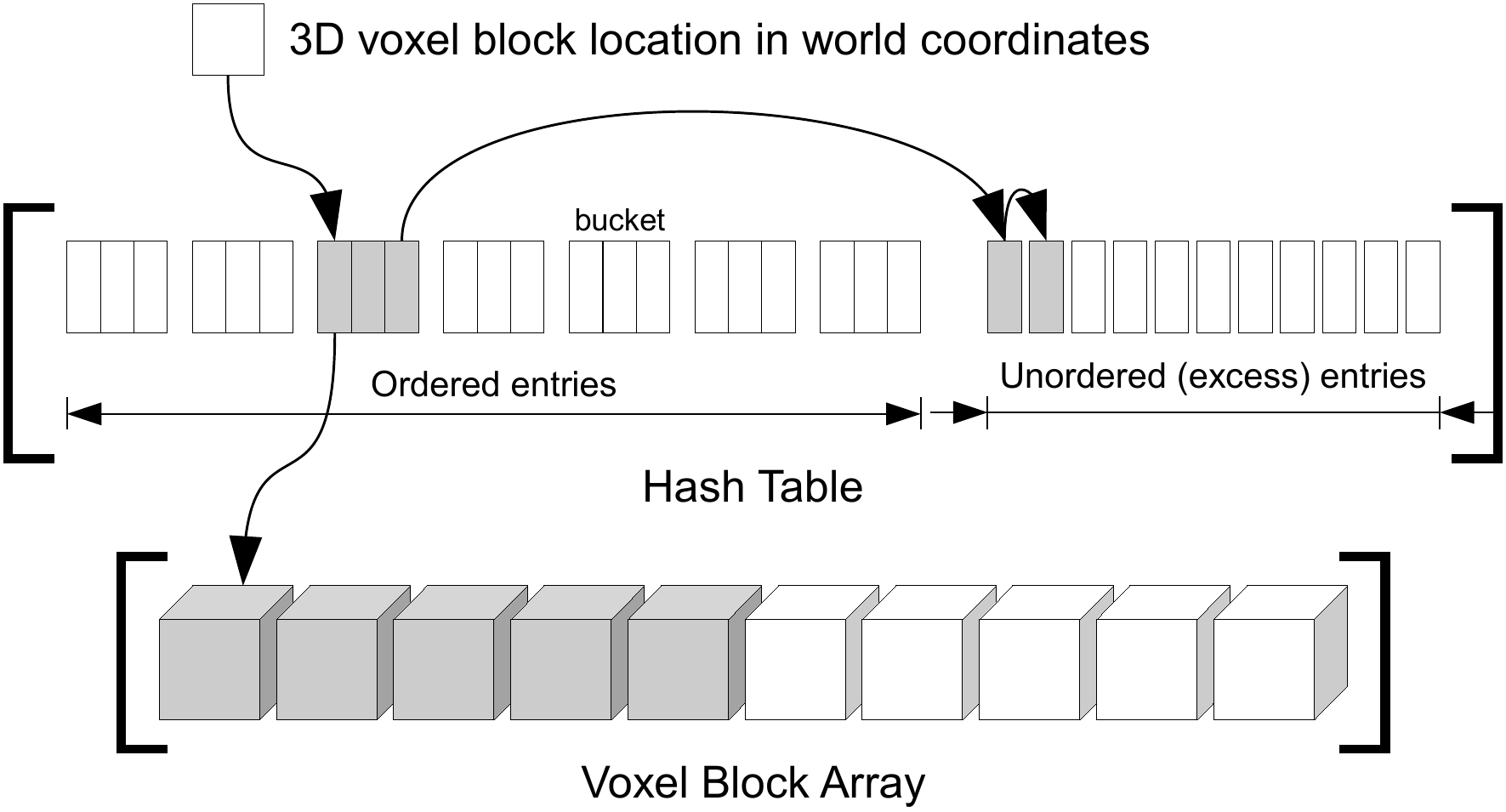}
\caption{Hash table logical structure}
\label{fig:itm_datastruct}
\end{figure}

To quickly and efficiently find the position of a certain voxel block in the
voxel block array, we use a hash table. This hash table is a contiguous array
of \reftocode{ITMHashEntry} objects of the following form:
\begin{lstlisting}
struct ITMHashEntry
{
	/** Position of the corner of the 8x8x8 volume, that identifies the entry. */
	Vector3s pos;
	/** Offset in the excess list. */
	int offset;
	/** Pointer to the voxel block array.
	    - >= 0 identifies an actual allocated entry in the voxel block array
	    - -1 identifies an entry that has been removed (swapped out)
	    - <-1 identifies an unallocated block
	*/
	int ptr;
};
\end{lstlisting}
The hash function \reftocode{hashIndex} for locating entries in the hash table
takes the corner coordinates \reftocode{vVoxPos} of a 3D voxel block and
computes an index as follows~\cite{neissner_tog_2013}:
\begin{lstlisting}
template<typename T> _CPU_AND_GPU_CODE_ inline int hashIndex(const InfiniTAM::Vector3<T> vVoxPos, const int hashMask){
	return ((uint)(((uint)vVoxPos.x * 73856093) ^ ((uint)vVoxPos.y * 19349669) ^ ((uint)vVoxPos.z * 83492791)) & (uint)hashMask);
}
\end{lstlisting}
To deal with \textit{hash collisions}, each hash index points to a
\textit{bucket} of fixed size (typically 2), which we consider the
\textit{ordered} part of the hash table. There is an additional
\textit{unordered} excess list that is used once an ordered bucket fills up.
In either case, each \reftocode{ITMHashEntry} in the hash table stores an offset
in the voxel block array and can hence be used to localise the voxel data for
this specific voxel block. This overall structure is illustrated in
Figure~\ref{fig:itm_datastruct}.

\subsubsection{Hash Table Operations} \label{ss:itm_hashopers}
The three main operations used when working with a hash table are the
\textbf{insertion}, \textbf{retrieval} and \textbf{removal} of entries. In the
current version of InfiniTAM we support the former two, with \textbf{removal}
not currently required or implemented. The code used by the \textbf{retrieval}
operation is shown below:
\begin{lstlisting}
template<class TVoxel>
_CPU_AND_GPU_CODE_ inline TVoxel readVoxel(const TVoxel *voxelData, const ITMVoxelBlockHash::IndexData *voxelIndex, const Vector3i & point, bool &isFound)
{
	const ITMHashEntry *hashTable = voxelIndex->entries_all;
	const TVoxel *localVBA = voxelData;
	TVoxel result; Vector3i blockPos; int offsetExcess;

	int linearIdx = vVoxPosParse(point, blockPos);
	int hashIdx = hashIndex(blockPos, SDF_HASH_MASK) * SDF_ENTRY_NUM_PER_BUCKET;

	isFound = false;

	//check ordered list
	for (int inBucketIdx = 0; inBucketIdx < SDF_ENTRY_NUM_PER_BUCKET; inBucketIdx++) 
	{
		const ITMHashEntry &hashEntry = hashTable[hashIdx + inBucketIdx];
		offsetExcess = hashEntry.offset - 1;

		if (hashEntry.pos == blockPos && hashEntry.ptr >= 0)
		{
			result = localVBA[(hashEntry.ptr * SDF_BLOCK_SIZE3) + linearIdx];
			isFound = true;
			return result;
		}
	}

	//check excess list
	while (offsetExcess >= 0)
	{
		const ITMHashEntry &hashEntry = hashTable[SDF_BUCKET_NUM * SDF_ENTRY_NUM_PER_BUCKET + offsetExcess];

		if (hashEntry.pos == blockPos && hashEntry.ptr >= 0)
		{
			result = localVBA[(hashEntry.ptr * SDF_BLOCK_SIZE3) + linearIdx];
			isFound = true;
			return result;
		}

		offsetExcess = hashEntry.offset - 1;
	}

	return result;
}
\end{lstlisting}

Both \textbf{insertion} and \textbf{retrieval} work by iterating through the
elements of the list stored within the hash table. Given a target 3D voxel
location in world coordinates, we first compute its corresponding voxel block
location, by dividing the voxel location by the size of the voxel blocks. Next,
we call the hashing function \reftocode{hashIndex} to compute the index of the
bucket from the ordered part of the hash table. All elements in the bucket are
then checked, with \textbf{retrieval} looking for the target block location and
\textbf{insertion} for an unallocated hash entry. If this is found,
\textbf{retrieval} returns the voxel stored at the target location within the
block addressed by the hash entry. \textbf{Insertion} (i) reserves a block
inside the voxel block array and (ii) populates the hash table with a new entry
containing the reserved voxel block array address and target block 3D world
coordinate location.

If all locations in the bucket are exhausted, the enumeration of the list moves
to the linked list in the unordered part of the hash table, using the
\reftocode{offset} field to provide the location of the next hash entry. The
enumeration finishes when \reftocode{offset} is found to be smaller or equal to
$-1$. At this point, if the target location still has not been found,
\textbf{retrieval} returns an empty voxel. \textbf{Insertion} (i) reserves an
unallocated entry in the unordered part of the hash table and a block inside
the voxel block array, (ii) populates the hash table with a new entry
containing the reserved voxel block array address and target block 3D world
coordinate location and (iii) changes the \reftocode{offset} field in the
previous entry in the linked list to point to the newly populated one.

The reserve operations used for the unordered part of the hash table and for
the voxel block array use prepopulated allocation lists and, in the GPU code,
atomic operations. 

All hash table operations are done through these functions and there is no
direct memory access encouraged or indeed permitted by the current version of
the code. 

\section{Individual Method Stages}\label{s:itm_methodstages}
\begin{figure}[tb]
\centering
\includegraphics[width=.9\linewidth]{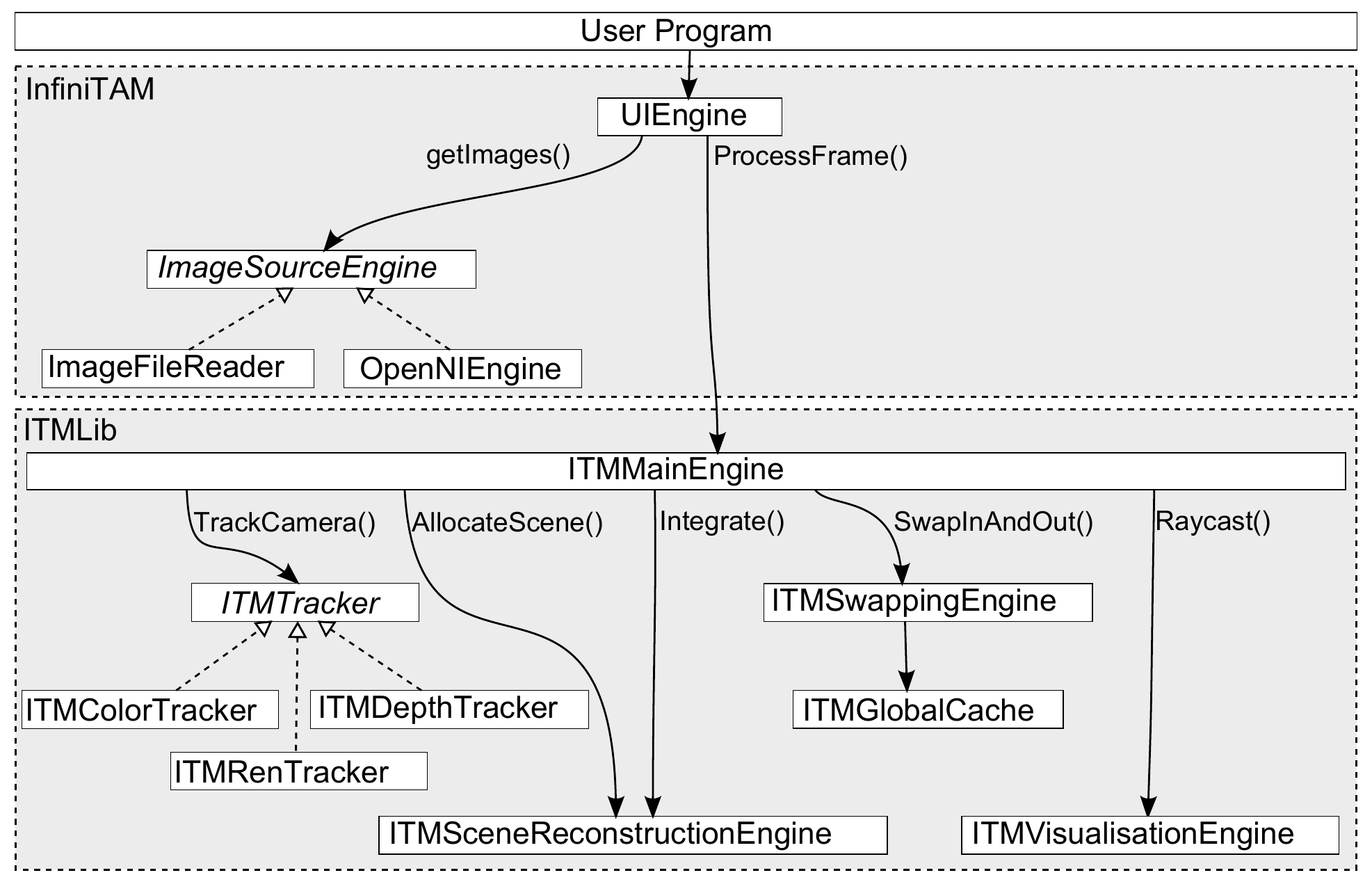}
\caption{Simplified diagram of InfiniTAM namespaces, classes and collaborations.}
\label{fig:itm_classdiagram}
\end{figure} 

A general outline of the InfiniTAM processing pipeline has already been given
in Section~\ref{s:pipeline} and Figure~\ref{fig:itm_pipeline}. Details of each
processing stage will be discussed in the following. The distinct stages are
implemented using individual engines and a simplified diagram of the
corresponding classes and their collaborations is given in
Figure~\ref{fig:itm_classdiagram}. The stages are:
\begin{itemize}
\item \textbf{Tracking}: The camera pose for the new frame is obtained by fitting the current depth (or optionally colour) image to the projection of the world model from the previous frame. This is implemented using the \reftocode{ITMTracker}, \reftocode{ITMColorTracker} and \reftocode{ITMDepthTracker} engines.
\item \textbf{Allocation}: Based on the depth image, new voxel blocks are allocated as required and a list of all visible voxel blocks is built. This is implemented inside the \reftocode{ITMSceneReconstructionEngine}.
\item \textbf{Integration}: The current depth and colour frames are integrated within the map. This is implemented inside the \reftocode{ITMSceneReconstructionEngine}.
\item \textbf{Swapping In and Out}: If required, map data is swapped in from host memory to device memory and merged with the present data. Parts of the map that are not required are swapped out from device memory to host memory. This is implemented using the \reftocode{ITMSwappingEngine} and \reftocode{ITMGlobalCache}.
\item \textbf{Raycasting}: The world model is rendered from the current pose (i) for visualisation purposes and (ii) to be used by the tracking stage at the next frame. This uses the \reftocode{ITMVisualisationEngine}.
\end{itemize}
As illustrated in Figure~\ref{fig:itm_classdiagram}, the main processing
engines are contained within the \reftocode{ITMLib} namespace.
Apart from these, the UI and image acquisition engines (\reftocode{UIEngine},
\reftocode{ImageSourceEngine} and \reftocode{OpenNIEngine}) are contained in
the \reftocode{InfiniTAM} namespace.
The \reftocode{ITMLib} namespace also contains the additional engine class
\reftocode{ITMLowLevelEngine} that we do not discuss in detail. It is used
for low level image processing such as computation of image gradients and a
resolution hierarchy.

In this section we discus the tracking, allocation, integration and raycasting
stages in greater detail. We delay a discussion of the swapping until
Section~\ref{s:swapping}.

\subsection{Tracking}

In the tracking stage the camera pose consisting of the rotation matrix
$\mathbf{R}$ and translation vector $\mathbf{t}$ has to be
determined given the RGB-D image and the current 3D world model. Along with
InfiniTAM we provide the
three engines \reftocode{ITMDepthTracker} and \reftocode{ITMRenTracker} that performs tracking based on the
new depth image, and \reftocode{ITMColorTracker} that is using the colour
image. All three of these implement the abstract \reftocode{ITMTracker} class and
have implementations running on the CPU and on CUDA.

In the \reftocode{ITMDepthTracker} we follow the original alignment process as
described in~\cite{newcombe_ismar_2011,izadi_uist_2011}:
\begin{itemize}
  \item Render a map $\mathcal{V}$ of surface points and a map $\mathcal{N}$ of
        surface normals from the viewpoint of an initial guess for $\mathbf{R}$
        and $\mathbf{t}$ -- details in Section~\ref{s:raycasting}
  \item Project all points $\mathbf{p}$ from the depth image onto points
	$\bar{\mathbf{p}}$ in $\mathcal{V}$ and $\mathcal{N}$ and compute their
        distances from the planar approximation of the surface,
        i.e.\ $d = \left(\mathbf{R} \mathbf{p} + \mathbf{t} - \mathcal{V}(\bar{\mathbf{p}})\right)^T \mathcal{N}(\bar{\mathbf{p}})$
  \item Find $\mathbf{R}$ and $\mathbf{t}$ minimising the linearised sum of the
        squared distances by solving a linear equation system
  \item Iterate the previous two steps until convergence
\end{itemize}
A resolution hierarchy of the depth image is used in our implementation to
improve the convergence behaviour.

\reftocode{ITMRenTracker} is used as the local refinement after \reftocode{ITMDepthTracker}, and we implemented a variation of the tracking algorithm described in \cite{Ren12:ECCV}:
\begin{itemize}
\item Given an initial guess for $\mathbf{R}$ and $\mathbf{t}$, the method back projects all points $\mathbf{p}$ from the depth image into points $\bar{\mathbf{p}}$ in the map $\mathcal{V}$ and computes the robust cost $\mathcal{E} = -4\exp\{\sigma\mathcal{V}(\bar{\mathbf{p}})\}/(\exp\{\sigma\mathcal{V}(\bar{\mathbf{p}})\} + 1)^2$, where $\sigma$ is a fixed parameter that controls the basin of attraction
\item Find $\mathbf{R}$ and $\mathbf{t}$ minimising the sum of the cost using Gauss Newton optimization algorithm
\end{itemize}
When \reftocode{ITMRenTracker} is chosen, the system first runs \reftocode{ITMDepthTracker} on coarse hierarchies of the depth image, then \reftocode{ITMRenTracker} runs on the high-resolution depth image for the final pose refinement. 

Alternatively the colour image can be used within an \reftocode{ITMColorTracker}.
In this case the alignment process is as follows:
\begin{itemize}
  \item Create a list $\mathcal{V}$ of surface points and a corresponding list
        $\mathcal{C}$ of colours from the viewpoint of an initial guess --
        details in Section~\ref{s:raycasting}
  \item Project all points from $\mathcal{V}$ into the current colour image $I$
        and compute the Euclidean norm of the difference in colours,
        i.e.\ $d = \left\|I(\pi(\mathbf{R} \mathcal{V}(i) + \mathbf{t})) - \mathcal{C}(i)\right\|_2$
  \item Find $\mathbf{R}$ and $\mathbf{t}$ minimising the sum of the squared
        differences using the Levenberg-Marquardt optimisation algorithm
\end{itemize}
Again a resolution hierarchy in the colour image is used and the list of surface
points is subsampled by a factor of 4. A flag in \reftocode{ITMLibSettings}
allows to select which tracker is used and the default is the
\reftocode{ITMDepthTracker}.

The three main classes \reftocode{ITMDepthTracker}, \reftocode{ITMRenTracker}  and \reftocode{ITMColorTracker}
actually only implement a shared optimisation framework, including e.g.\ the
Levenberg-Marquardt algorithm, Gauss Newton algorithm and solving the linear equation systems. These
are always running on the CPU. Meanwhile the evaluation of the error function
value, gradient and Hessian is implemented in derived, CPU and CUDA specific
classes and makes use of parallelisation.

\subsection{Allocation}
In the allocation stage the underlying data structure for the representation
of the truncated signed distance function is prepared and updated, so that a new
depth image can be integrated. In the simple case of a dense voxel grid, the
allocation stage does nothing. In contrast, for voxel block hashing the goal
is to find all the voxel blocks affected by the current depth image and to
make sure that they are allocated in the hash table~\cite{neissner_tog_2013}.

In our implementation of voxel block hashing, the aim was to minimise the use
of blocking operations (e.g.\ atomics) and to completely avoid the use of
critical sections. This has led us to doing the processing three separate
stages, as we explain in the following.

In the first stage we project each pixel from the depth image to 3D space
and create a line segment along the ray from depth $d-\mu$ to $d+\mu$, where
$d$ is the measured depth at the pixel and $\mu$ is the width of the truncation
band of the signed distance function. This line segment intersects a number of
voxel blocks, and we search for these voxel blocks in the hash table. If one of
the blocks is not allocated yet, we find a free hash bucket space for it. As a
result for the next stage we create two arrays, each of the same size as the
number of elements in the hash table. The first array contains a bool
indicating the visibility of the voxel block referenced by the hash table entry,
the second contains information about new allocations that have to be performed.
Note that this requires only
simple, non-atomic writes and if more than one new block has to be allocated
with the same hash index, only the most recently written allocation will
actually be performed. We tolerate such artefacts from intra-frame hash
collisions, as they will be corrected in the next frame automatically for
small intra-frame camera motions.

In the second stage we allocate voxel blocks for each non-zero entry in the
allocation array that we built previously. This is done using a single atomic
subtraction on a stack of free voxel block indices i.e.\ we decrease the number
of remaining blocks by one and add the previous head of the stack to the hash
entry.

In the third stage we build a list of live voxel blocks, i.e.\ a list of the
blocks that project inside the visible view frustum. This is later going to be
used by the integration and swapping stages. 

\subsection{Integration}

In the integration stage, the information from the most recent RGB-D image is
incorporated into the 3D world model. In case of a dense voxel grid this is
identical to the integration in the original KinectFusion
algorithm~\cite{newcombe_ismar_2011,izadi_uist_2011} and for voxel block
hashing the changes are minimal after the list of visible voxel blocks has been
created in the allocation step.

For each voxel in any of the visible voxel blocks, or for each voxel in the
whole volume for dense grids, the function
\reftocode{computeUpdatedVoxelDepthInfo} is called. If a voxel is
\textit{behind} the surface observed in the new depth image by more than the
truncation band of the signed distance function, the image does not contain any
new information about this voxel, and the function returns without writing
any changes. If the voxel is close to or in front of the observed surface, a
corresponding observation is added to the accumulated sum. This is illustrated
in the listing of the function \reftocode{computeUpdatedVoxelDepthInfo} below,
and there is a similar function in the code that additionally updates the colour
information of the voxels.
\begin{lstlisting}
template<class TVoxel>
_CPU_AND_GPU_CODE_ inline float computeUpdatedVoxelDepthInfo(TVoxel &voxel, Vector4f pt_model, Matrix4f M_d, Vector4f projParams_d,
	float mu, int maxW, float *depth, Vector2i imgSize)
{
	Vector4f pt_camera; Vector2f pt_image;
	float depth_measure, eta, oldF, newF;
	int oldW, newW;

	// project point into image
	pt_camera = M_d * pt_model;
	if (pt_camera.z <= 0) return -1;

	pt_image.x = projParams_d.x * pt_camera.x / pt_camera.z + projParams_d.z;
	pt_image.y = projParams_d.y * pt_camera.y / pt_camera.z + projParams_d.w;
	if ((pt_image.x < 1) || (pt_image.x > imgSize.x - 2) || (pt_image.y < 1) || (pt_image.y > imgSize.y - 2)) return - 1;

	// get measured depth from image
	depth_measure = depth[(int)(pt_image.x + 0.5f) + (int)(pt_image.y + 0.5f) * imgSize.x];
	if (depth_measure <= 0.0) return -1;

	// check whether voxel needs updating
	eta = depth_measure - pt_camera.z;
	if (eta < -mu) return eta;

	// compute updated SDF value and reliability
	oldF = TVoxel::SDF_valueToFloat(voxel.sdf); oldW = voxel.w_depth;
	newF = MIN(1.0f, eta / mu);
	newW = 1;

	newF = oldW * oldF + newW * newF;
	newW = oldW + newW;
	newF /= newW;
	newW = MIN(newW, maxW);

	// write back
	voxel.sdf = TVoxel::SDF_floatToValue(newF);
	voxel.w_depth = newW;

	return eta;
}
\end{lstlisting}
The main difference between the dense voxel grid and the voxel block hashing
representations is that the aforementioned update function is called for
a different number of voxels and from within different loop constructs.

\subsection{Raycast}\label{s:raycasting}
As the last step in the pipeline, an image is computed from the updated 3D
world model to provide input for the tracking at the next frame. This image can
also be used for visualisation. The main process underlying this rendering is
raycasting, i.e.\ for each pixel in the image a ray is being cast from the
camera up until an intersection with the surface is found. This essentially
means checking the value of the truncated signed distance function at each
voxel along the ray until a zero-crossing is found, and the same raycasting
engine can be used for a range of different representations, as long as an
appropriate \reftocode{readVoxel()} function is called for reading values from
the SDF.

As noted in the original KinectFusion paper~\cite{newcombe_ismar_2011}, the
performance of the raycasting can be improved significantly by taking larger
steps along the ray. The value of the truncated signed distance function can
serve as a conservative estimate for the distance to the nearest surface, hence
this value can be used as step size. To additionally handle empty space in the
volumetric representation, where no corresponding voxel block has been
allocated, we introduce a state machine with the following states:
\begin{lstlisting}
enum {
	SEARCH_BLOCK_COARSE,
	SEARCH_BLOCK_FINE,
	SEARCH_SURFACE,
	BEHIND_SURFACE,
	WRONG_SIDE
} state;
\end{lstlisting}
Starting from \reftocode{SEARCH\_BLOCK\_COARSE}, we take steps of the size of
each block, i.e.\ $8$ voxels, until an actually allocated block is encountered.
Once the ray enters an allocated block, we take a step back and enter state
\reftocode{SEARCH\_BLOCK\_FINE}, indicating that the step length is now limited
by the truncation band $\mu$ of the signed distance function. Once we enter a
valid block and the values in that block indicate we are still in front of the
surface, the state is changed to \reftocode{SEARCH\_SURFACE} until a negative
value is read from the signed distance function, which indicates we are now
in state \reftocode{BEHIND\_SURFACE}. This terminates the raycasting iteration
and the exact location of the surface is now found using two trilinear
interpolation steps. The state \reftocode{WRONG\_SIDE} is entered if we are
searching for a valid block in state \reftocode{SEARCH\_BLOCK\_FINE} and
encounter negative SDF values, indicating we are behind the surface as soon as
we enter a block. In this case the ray hits the surface from behind for
whichever reason, and we do not want to count the boundary between the
unallocated, empty space and the block with the negative values as an object
surface.

Another measure for improving the performance of the raycasting is to select a
plausible search range. If a sparse voxel block representation is used, then
we are given a list of visible blocks from the allocation step, and we can
render these blocks by forward projection to give us an idea of the maximum and
minimum depth values to expect at each pixel. Within InfiniTAM this can be done
using the method \reftocode{CreateExpectedDepths()} of an
\reftocode{ITMVisualisationEngine}. A naive implementation on the
CPU computes the 2D bounding box of the projection of each voxel block into the
image and fills this area with the maximum and minimum depth values of the
corresponding 3D bounding box of the voxel block, correctly handling
overlapping bounding boxes, of course.

To parallelise this process on the GPU we split it into two steps. First we
project each block down into the image, compute the bounding box, and create a
list of $16\times{}16$ pixel fragments, that are to be filled with specific
minimum and maximum depth values. Apart from a prefix sum to count the number
of fragments, this is trivially parallelisable. Second we go through the list
of fragments and actually render them. Updating the minimum and maximum depth
for a pixel requires atomic operations, but by splitting the process into
fragments we reduce the number of collisions to typically a few hundreds or
thousands of pixels in a $640\times{}480$ image and achieve an efficiently
parallelised overall process.

\section{Swapping}\label{s:swapping}
Voxel hashing already enables much larger maps to be created, compared to the much simpler dense 3D volumes. Video card memory capacity however is often quite limited. Practically an off-the-shelf video card can roughly hold the map of a single room at 4mm voxel resolution in active memory, even with voxel hashing. This problem can be mitigated using a traditional method from the graphics community, that is also employed e.g.\ in~\cite{kinfuls_2012,Chen13:SRV,neissner_tog_2013,Henry13:PVS}. We only hold the \textit{active} part of the map in video card memory, i.e.\ only parts that are inside or close to the current view frustum. The remainder of the map is swapped out to host memory and swapped back in as needed. 

We have designed our swapping framework aiming for the following three objectives: (O1) the transfers between host and device should be minimised and have guaranteed maximum bounds, (O2) host processing time should be kept to a minimum and (O3) no assumptions should be made about the type and speed of the host memory, i.e.\ it could be a hard drive. These objectives lead to the following design considerations:
\begin{itemize}
\item \textbf{O1}: All memory transfers use a host/device buffer of \textit{fixed} user-defined size. 
\item \textbf{O2}: The host map memory is configured as a voxel block array of size equal to the number of entries in the hash table. Therefore, to check if a hash entry has a corresponding voxel block in the host memory, only the hash table index needs to be transferred and checked. The host does not need to perform any further computations, e.g.\ as it would have to do if a separate host hash table were used. Furthermore, whenever a voxel block is deallocated from device memory, its corresponding hash entry is not deleted but rather marked as unavailable in device memory, and, implicitly, available in host memory. This (i) helps maintain consistency between device hash table and host voxel block storage and (ii) enables a fast visibility check for the parts of the map stored only in host memory.
\item \textbf{O3}: Global memory speed is a function of the type of storage device used, e.g.\ faster for RAM and slower for flash or hard drive storage. This means that, for certain configurations, host memory operations can be considerably slower than the device reconstruction. To account for this behaviour and to enable stable tracking, the device is constantly integrating new live depth data even for parts of the scene that are known to have host data that is not yet in device memory. This might mean that, by the time all visible parts of the scene have been swapped into the device memory, some voxel blocks might hold large amounts of new data integrated by the device. We could replace the newly fused data with the old one from the host stored map, but this would mean disregarding perfectly fine map data. Instead, after the swapping operation, we run a secondary integration that fuses the host voxel block data with the newly fused device map.
\end{itemize}

\begin{figure}[tb]
\centering
\includegraphics[width=0.7\linewidth]{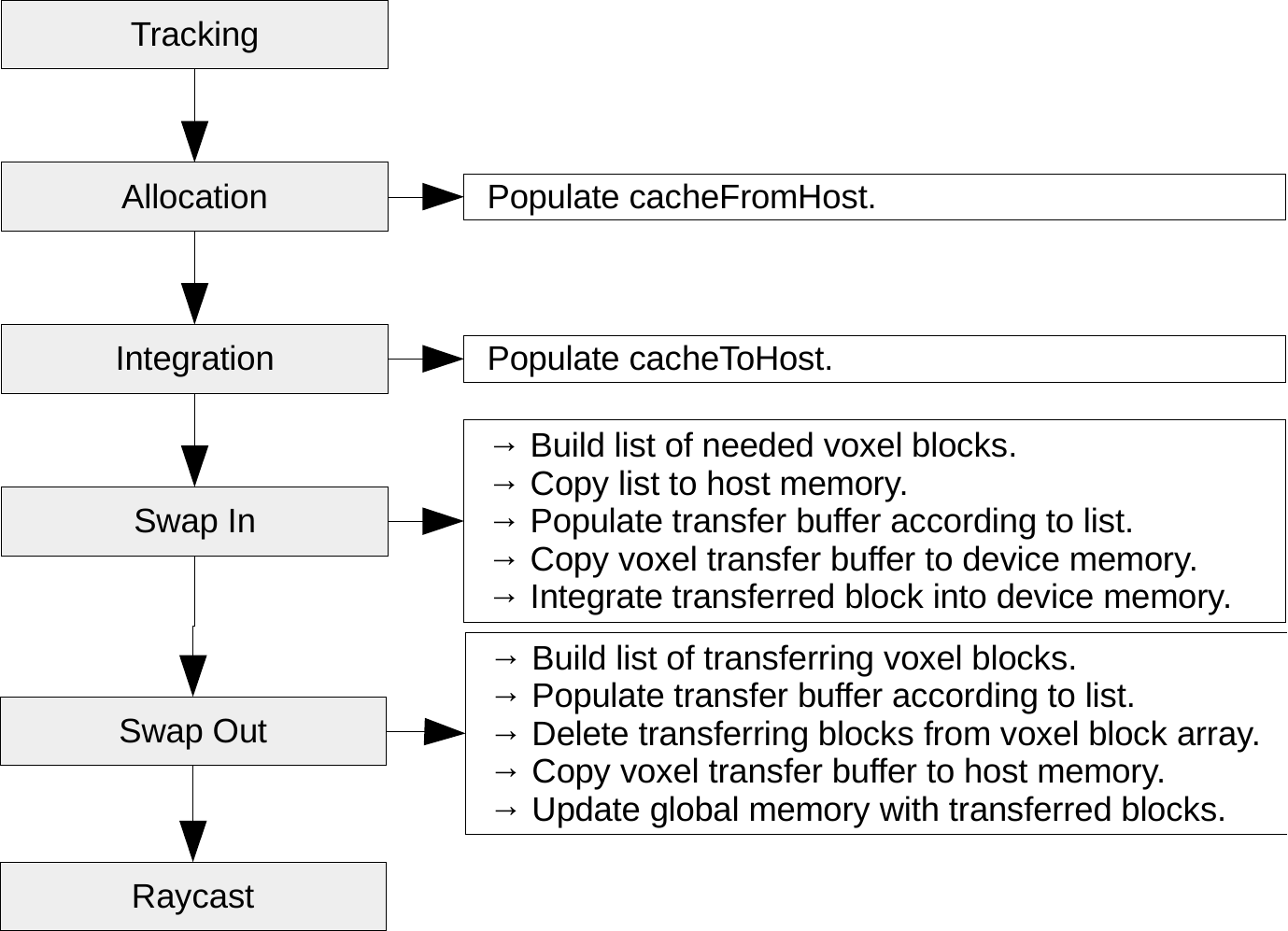}
\caption{Swapping pipeline}
\label{fig:itm_swap_overall}
\end{figure} 

The design considerations have led us to the swapping in/out pipeline shown in Figure~\ref{fig:itm_swap_overall}. We use the allocation stage to establish which parts of the map need to be swapped in, and the integration stage to mark which parts need to swapped out. A voxel needs to be swapped (i) from host once it projects within a small (tunable) distance from the boundaries of live visible frame and (ii) to disk after data has been integrated from the depth camera. 

\begin{figure}[tb]
\centering
\includegraphics[width=0.9\linewidth]{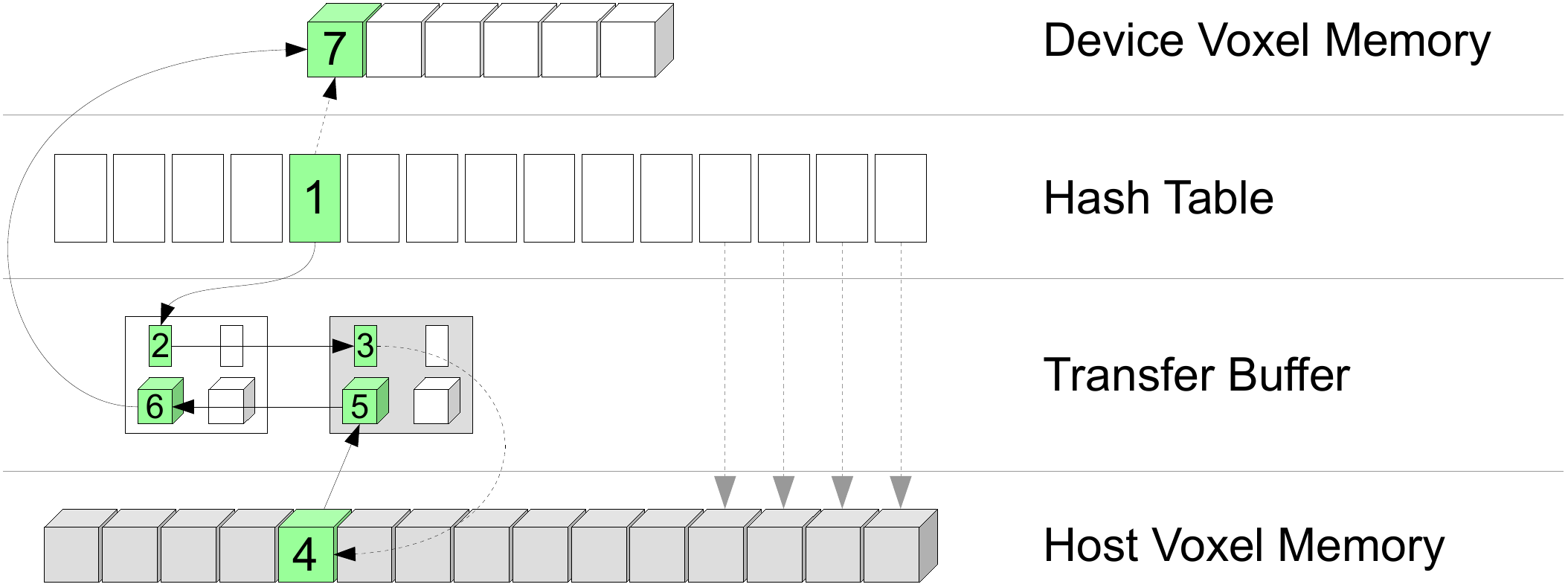}
\caption{Swapping in: First, the hash table entry at address \textbf{1} is copied into the device transfer buffer at address \textbf{2}. This is then copied at address \textbf{3} in the host transfer buffer and used as an address inside the host voxel block array, indicating the block at address \textbf{4}. This block is finally copied back to location \textbf{7} inside the device voxel block array, passing through the host transfer buffer (location \textbf{5}) and the device transfer buffer (location \textbf{6}).}
\label{fig:itm_swap_in}
\end{figure} 

The swapping in stage is exemplified for a single block in Figure \ref{fig:itm_swap_in}. The indices of the hash entries that need to be swapped in are copied into the device transfer buffer, up to its capacity. Next, this is transferred to the host transfer buffer. There the indices are used as addresses inside the host voxel block array and the target blocks are copied to the host transfer buffer. Finally, the host transfer buffer is copied to the device where a single kernel integrates directly from the transfer buffer into the device voxel block memory.

\begin{figure}[tb]
\centering
\includegraphics[width=0.9\linewidth]{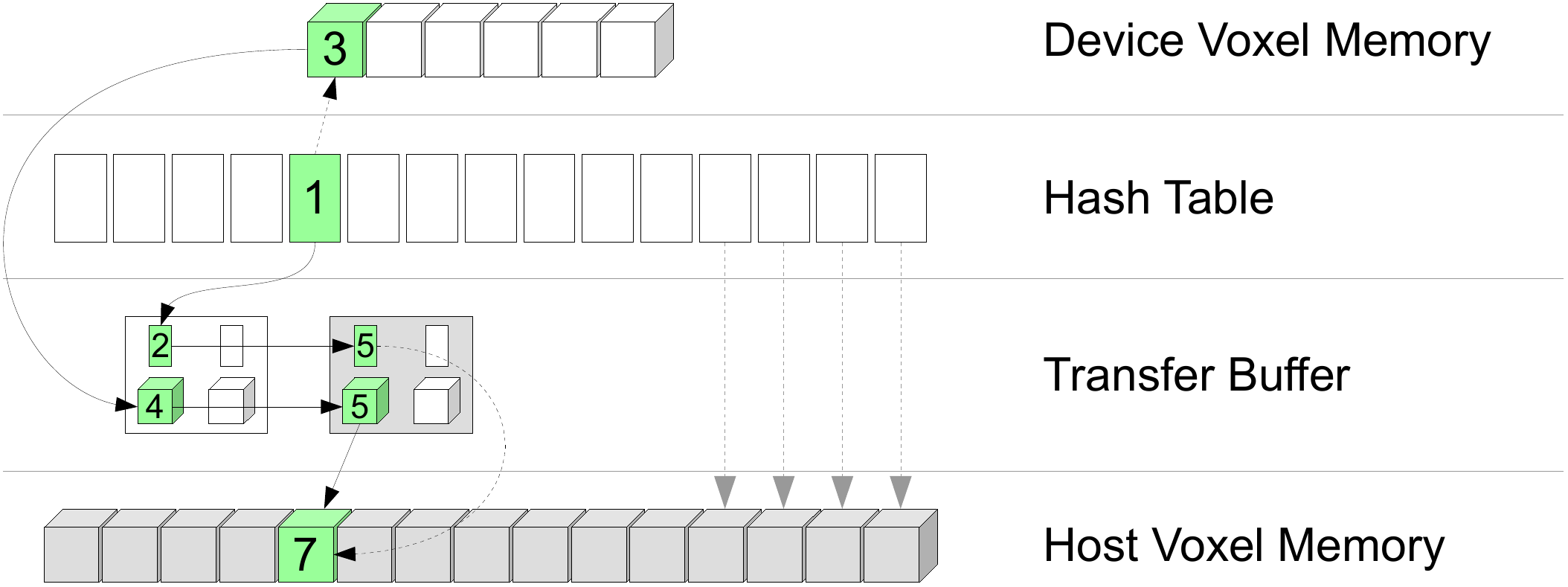}
\caption{Swapping Out. The hash table entry at location \textbf{1} and the voxel block at location \textbf{3} are copied into the device transfer buffer at locations \textbf{2} and \textbf{4}, respectively. The entire used transfer buffer is then copied to the host, at locations \textbf{5}, and the hash entry index is used to copy the voxel block into location \textbf{7} inside the host voxel block array.}
\label{fig:itm_swap_out}
\end{figure} 

An example for the swapping out stage is shown in Figure \ref{fig:itm_swap_out} for a single block. Both indices and voxel blocks that need to be swapped out are copied to the device transfer buffer. This is then copied to the host transfer buffer memory and again to host voxel memory.

All swapping related variables and memory is kept inside the \reftocode{ITMGlobalCache} object and all swapping related operations are done by the \reftocode{ITMSwappingEngine}. 

\section{Compilation, UI, Usage and Examples}\label{s:results}
As indicated in the class diagram in Figure~\ref{fig:itm_classdiagram}, the
InfiniTAM software package is split into two major parts. The bulk of the
implementation is grouped in the namespace \reftocode{ITMLib}, which contains
a stand alone 3D reconstruction library for use in other applications.
The \reftocode{InfiniTAM} namespace contains further supporting classes for
image acquisition and a GUI -- these parts would almost certainly be replaced
in user applications. Finally there is a single sample application included
allowing the user to quickly test the framework.

The project comes with a Microsoft\textregistered{} Visual
Studio\textregistered{} solution file as well as with a cmake build file and
has been tested on Microsoft\textregistered{} Windows\textregistered{} 8,
openSUSE Linux 12.3 and Apple\textregistered{} Mac\textregistered{} OS X\textregistered{} 10.9 platforms. Apart from a basic C++
software development environment, InfiniTAM depends on the following external
third party libraries:
\begin{itemize}
  \item \textbf{OpenGL / GLUT} (e.g.\ freeglut 2.8.0): This is required for
        the visualisation in the \reftocode{InfiniTAM} namespace and the
        sample application, but the \reftocode{ITMLib} library should run
        without. Freeglut is available from
        \url{http://freeglut.sourceforge.net/}
  \item \textbf{NVIDIA\textcopyright{} CUDA\texttrademark{} SDK}
        (e.g.\ version 6.0): This is required for all GPU accelerated code.
        The use of GPUs is optional however, and it is still possible to
        compile the CPU part of the framework without CUDA. The CUDA SDK is
        available from \url{https://developer.nvidia.com/cuda-downloads}
  \item \textbf{OpenNI} (e.g.\ version 2.2.0.33): This is optional and the
        framework compiles without OpenNI, but it is required to get live
        images from suitable hardware sensors. Again, it is only referred to
        in the \reftocode{InfiniTAM} namespace, and without OpenNI the
        system will still run of previously recorded images stored on disk.
        OpenNI is available from \url{http://structure.io/openni}
\end{itemize}
Finally the framework comes with a Doxygen reference documentation, that can
be built separately. More details on the build process can be found in the
\texttt{README} file provided alongside the framework.

\begin{figure}[tb]
\centering
\includegraphics[width=0.9\linewidth]{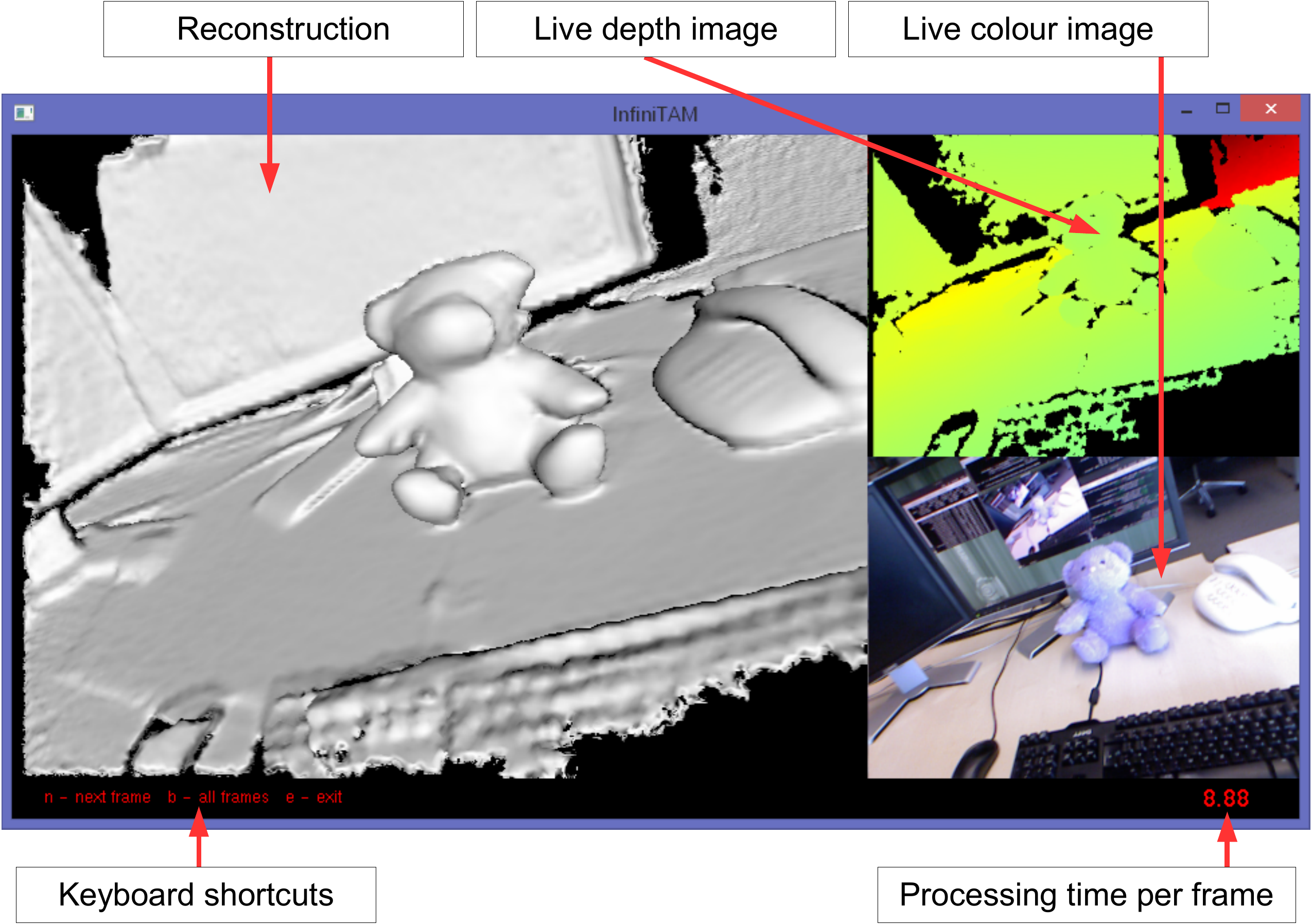}
\caption{InfiniTAM UI Example}
\label{fig:itm_ui}
\end{figure} 

The UI for our InfiniTAM sample application is shown in Figure~\ref{fig:itm_ui}.
We display the raycasted reconstruction rendering, live depth image and live
colour image from the camera. Furthermore the processing time per frame and the
keyboard shortcuts available for the UI are displayed near the bottom. The
keyboard shortcuts allow the user to process the next frame, to process
continuously and to exit. Other functionalities such as exporting a 3D model
from the current reconstruction and rendering the reconstruction from arbitrary
viewpoints are not currently implemented. The UI window and interactivity is
implemented in \reftocode{UIEngine} and depends on the OpenGL and GLUT
libraries.

Running InfiniTAM requires the intrinsic calibration data for the depth (and
optionally colour) camera. The calibration is specified through an external
file, an example of which is shown below.
\begin{lstlisting}
640 480
504.261 503.905
352.457 272.202

640 480
573.71 574.394
346.471 249.031

0.999749 0.00518867 0.0217975 0.0243073
-0.0051649 0.999986 -0.0011465 -0.000166518
-0.0218031 0.00103363 0.999762 0.0151706

1135.09 0.0819141
\end{lstlisting}
This includes (i) for each camera (RGB and depth) the image size, focal length
and principal point in pixels, as per the Matlab Calibration
Toolbox~\cite{Bouguet04:CCT} (ii) the Euclidean transformation matrix mapping
points in the RGB camera coordinate system to the depth camera
and (iii) the calibration converting Kinect-like disparity values to depths.
If the depth tracker is used, the
calibration for the RGB camera and the Euclidean transformation between the two
are ignored. If live data from OpenNI is used, the disparity calibration is
ignored.

We also provide two example command lines using different data sources, OpenNI
and image files. These data sources are selected based on the number of
arguments passed to InfiniTAM and for example in the bash shell with a
Linux environment these are:
\begin{small}
\begin{verbatim}
  $ ./InfiniTAM Teddy/calib.txt
  $ ./InfiniTAM Teddy/calib.txt Teddy/Frames/%04i.ppm Teddy/Frames/%04i.pgm
\end{verbatim}
\end{small}
The first line starts InfiniTAM with the specified calibration file and live
input from OpenNI, while the second uses the given calibration file and the
RGB and depth images specified in the other two arguments.
We tested the OpenNI input with a Microsoft Kinect for XBOX 360, with a
PrimeSense Carmine 1.08 
and with the Occipital Structure Sensor. 
For offline use with images from files, these have to be in PPM/PGM format
with the RGB images being standard PPM files and the depth images being 16bit
big-endian raw Kinect disparities.

All internal library settings are defined inside the \reftocode{ITMLibSettings}
class, and they are:
\begin{lstlisting}
/// Use GPU or run the code on the CPU instead.
bool useGPU;

/// Enables swapping between host and device.
bool useSwapping;

/// Tracker types
typedef enum {
	//! Identifies a tracker based on colour image
	TRACKER_COLOR,
	//! Identifies a tracker based on depth image
	TRACKER_ICP
} TrackerType;
/// Select the type of tracker to use
TrackerType trackerType;

/// Number of resolution levels for the tracker.
int noHierarchyLevels;

/// Number of resolution levels to track only rotation instead of full SE3.
int noRotationOnlyLevels;
			
/// For ITMColorTracker: skip every other point in energy function evaluation.
bool skipPoints;

/// For ITMDepthTracker: ICP distance threshold
float depthTrackerICPThreshold;

/// Further, scene specific parameters such as voxel size
ITMLib::Objects::ITMSceneParams sceneParams;
\end{lstlisting}
The \reftocode{ITMSceneParams} further contain settings for the voxel size in
millimetres and the truncation band of the signed distance function.
Furthermore the file \reftocode{ITMLibDefines.h} contains definitions that
select the type of voxels and the voxel index used in the compilation of
\reftocode{ITMMainEngine}:
\begin{lstlisting}
/** This chooses the information stored at each voxel. At the moment, valid
    options are ITMVoxel_s, ITMVoxel_f, ITMVoxel_s_rgb and ITMVoxel_f_rgb 
*/
typedef ITMVoxel_s ITMVoxel;

/** This chooses the way the voxels are addressed and indexed. At the moment,
    valid options are ITMVoxelBlockHash and ITMPlainVoxelArray.
*/
typedef ITMLib::Objects::ITMVoxelBlockHash ITMVoxelIndex;
\end{lstlisting}

For using InfiniTAM as a 3D reconstruction library in other applications, the
class \reftocode{ITMMainEngine} is recommended as the main entry point to the
\reftocode{ITMLib} library. It performs the whole 3D reconstruction algorithm.
Internally it stores the latest image as well as the 3D world model and it
keeps track of the camera pose. The intended use is as follows:
\begin{enumerate}
  \item Create an \reftocode{ITMMainEngine} specifying the internal settings,
        camera parameters and image sizes.
  \item Get the pointer to the internally stored images with the method
        \reftocode{GetView()} and write new image information to that memory.
  \item Call the method \reftocode{ProcessFrame()} to track the camera and
        integrate the new information into the world model.
  \item Optionally access the rendered reconstruction or another image for
        visualisation using \reftocode{GetImage()}.
  \item Iterate the above three steps for each image in the sequence.
\end{enumerate}
The internally stored information can be accessed through member variables
\reftocode{trackingState} and \reftocode{scene}. 

\section{Conclusions}\label{s:conclusions}
We tried to keep the InfiniTAM system as simple, portable and usable as
possible. We hope that the external dependencies on CUDA, OpenNI, OpenGL and
GLUT are easy to meet and that the cmake and MSVC project files allow users to
build the framework on a wide range of platforms. While the user interface and
surroundings are fairly minimalistic, the underlying library should be reliable,
robust and well tested.

We also tried to keep an eye on expandability. For example it
is fairly easy to integrate a different form of tracking into the
pipeline by reimplementing the \reftocode{ITMTracker} interface. Also different
media for swapping can be made accessible by replacing the
\reftocode{ITMGlobalCache} class. Finally it is still a manageable
process to port InfiniTAM to different kinds of hardware platforms by
reimplementing the \textit{Device Specific Layer} of the engines.

While we hope that InfiniTAM provides a reliable basis for further research,
no system is perfect. We will try to fix any problems we hear of and provide
future updates to InfiniTAM at the project website.

\small
\bibliographystyle{plain}
\bibliography{InfiniTAM}

\end{document}